\documentclass[12pt]{article}
\usepackage{amsmath,amsfonts}
\usepackage{algorithm}
\usepackage{array}
\usepackage{textcomp}
\usepackage{stfloats}
\usepackage{url}
\usepackage{verbatim}
\usepackage{graphicx}
\usepackage{cite}
\usepackage{amsmath,geometry}
\usepackage{natbib}
\usepackage{algpseudocode}
\usepackage{lscape}
\usepackage{setspace}
\usepackage{authblk}
\usepackage{graphics,graphicx}
\usepackage{xcolor}
\usepackage{bm}
\usepackage{booktabs}
\usepackage{comment}
\usepackage{breqn}
\usepackage{booktabs}
\usepackage{hyperref}

\newcommand{\alphavec}{\boldsymbol{\alpha}}

\newcommand{\xvec}{\textbf{x}}

\date{}
\begin{document} 

\def\spacingset#1{\renewcommand{\baselinestretch}%
{#1}\small\normalsize} \spacingset{1}

\title{Boundary Peeling: An Outlier Detection Method }

\author[1]{\small Sheikh Arafat}
\author[1]{\small Na Sun}
\author[2]{\small Maria L. Weese}
\author[2]{\small Waldyn G. Martinez}
\affil[1]{\small Department of Statistics, Miami University, Oxford, OH}
\affil[2]{\small Department of Information Systems \& Analytics, Miami University, Oxford, OH}



\maketitle
\doublespace

\begin{abstract}

Unsupervised outlier detection constitutes a crucial phase within data analysis and remains an open area of research. A good outlier detection algorithm  should be computationally efficient, robust to tuning parameter selection, and perform consistently well across diverse underlying data distributions. We introduce Boundary Peeling, an unsupervised outlier detection algorithm. Boundary Peeling uses the average signed distance from iteratively-peeled, flexible boundaries generated by one-class support vector machines to flag outliers.  The method is similar to convex hull peeling but well suited for high-dimensional data and has flexibility to adapt to different distributions.  Boundary Peeling has robust hyperparameter settings and, for increased flexibility, can be cast as an ensemble method. In unimodal and multi-modal synthetic data simulations Boundary Peeling outperforms all state of the art methods when no outliers are present while maintaining comparable or superior performance in the presence of outliers. Boundary Peeling performs competitively or better in terms of correct classification, AUC, and processing time using semantically meaningful benchmark data sets.

\end{abstract}

\noindent%


\section{Introduction}

{Outlier detection is crucial in applications such as fraud detection \citep{westerski2021explainable}, medicine \citep{zhang2022unsupervised}, and network intrusion \citep{ponmalar2022intrusion}. Despite extensive research on outlier detection methods, there is no consensus on a single best approach \citep{aggarwal2017introduction}. This lack of consensus highlights the need for continual improvement and development of new methods.  While outliers have many definitions, they are generally considered objects that deviate significantly from most observations, suggesting they are generated by a different process \citep{hawkins1980identification}.}

 {Data-depth methods, such as convex hull peeling, have shown promise in the statistics literature for their distribution-free properties and flexibility to adapt to different distributions \citep{ruts1996computing, bell2014distribution}. However, a major limitation of these methods is their infeasibility for data with dimension $p > 9$ \citep{knox1998algorithms}. In this paper, we introduce an unsupervised method for outlier detection, similar to convex hull peeling, that uses the average signed distance from iteratively peeled, flexible boundaries generated by one-class support vector machines (OCSVM) \citep{scholkopf2001estimating}. Our method, Boundary Peeling (BP), offers robust default hyperparameters (unlike \(k\) in nearest neighbor methods) and performs well with a simple threshold for outlier identification. We demonstrate that the proposed method  has comparable computational performance and performance with many current state-of-the-art methods on benchmark and synthetic datasets. BP particularly excels when there are no outliers present, which is particularly appealing when the sample size \(N\) is small. Unlike covariance estimation-based outlier detection methods \citep{filzmoser2008outlier}, BP is feasible when the dimension \(p\) is smaller than \(N\) since it does not require the estimation of a covariance matrix. Boundary Peeling performs well regardless of the number of modes and with various percentages of outliers. For multimodal data, BP requires no pre-specification of the number of modes.}

{\cite{Martinez2020OCP} introduced One Class Peeling (OCP), a proximity-based method, for outlier detection using Support Vector Data Description (SVDD) \citep{tax1999support}. The OCP method involves fitting a SVDD boundary and then removing the points identified as support vectors, similar to a data-depth or a convex hull peeling algorithm. The peeling process is repeated until a small number, say 2, points remain.  These final points are used to estimate the multivariate center of data.  From that center, Gaussian kernel distances 
used to calculate a distance of each observation from the estimated center and large distances are flagged as outliers.  OCP is not dissimilar from convex hull peeling \citep{barnett1976ordering} but unlike convex hull peeling OCP is scaleable and flexible \citep{Martinez2020OCP}. The OCP method works reasonably well, but only on data with a single mode and when the user can pre-specify the underlying data distribution. While \cite{Martinez2020OCP} do provide empirical threshold values to control the false positive rate, those values are highly dependent on the data itself and simulations have shown the performance of OCP is highly dependent on the ability to choose the threshold for identification accurately. This is not a desirable property for a unsupervised outlier detection method.  Boundary Peeling makes the following improvements to OCP:}

\begin{itemize}
\item { \cite{lee2022robustness} found bias in the estimation of the center using the OCP. Boundary peeling works in a single step, not requiring estimation of a center and  $n$ distance calculations. Boundary Peeling  assigns signed distances to observations in relation to the successive boundaries themselves.}
    \item  {\cite{Martinez2020OCP} showed the performance of OCP to be highly dependent on the underlying distribution and requires an empirical threshold calculated based on the underlying distribution. Boundary Peeling performance is consistent across different distributions and uses only a single threshold value.}
    \item   {OCP will only work on data if it has a single mode.  Boundary Peeling will work on data, regardless of the number of modes.  Additionally there is no need to specify the number of modes ahead of time. }
    
\end{itemize}

 {Real-world data might be high dimensional and have distributions with multiple modes.  Multiple modes in data are generated from processes that operate under different modalities \citep{nedelkoski2019anomaly, sipple2020interpretable, chan2020ensemble, park2016multimodal, feng2022unsupervised}. Examples of multimodal data can be found in chemical engineering, environmental science, biomedicine, and the pharmaceutical business \citep{lahat2015multimodal,chiang2017big}.  In fact, all business processes that involve transitions or depend on time might be categorized as multimodal. According to \cite{quinones2019data}, a multimode process is one that operates properly at various operational points. It is important to identify and separate defects from modes in order to preserve data quality. In this work we assess the performance of our method, as well as others, in  unimodal and multimodal settings.  Ideally, a method should work well regardless of the number of modes and not require the specification of the number of modes \emph{a priori}.}

{Outlier detection methods can be supervised, semi-supervised, or unsupervised. Unsupervised outlier detection is the most challenging, but most realistic case, as labeled data is often unavailable.  Unsupervised outlier detection methods include ensemble-based, probabilistic methods, proximity-based methods, deep learning methods, and graph-based methods \citep{domingues2018comparative}.  An ideal method will scale to problems with high-dimensional and/or large data, perform regardless of the shape of the data, consistently identifies outliers in uni- or multimodal data, have minimal hyperparameter tuning, and be computationally efficient. A popular ensemble method that checks many of those boxes is the Isolation Forest \citep{liu2012isolation}. }

{The Isolation Forest (ISO) algorithm identifies outliers by generating recursive random splits on attribute values which isolate anomalous observations using the path length from the root node. 
The popularity of ISO is due its accuracy, flexibility, and its linear time complexity.  \cite{domingues2018comparative} show the ISO to be the preferable method among other unsupervised methods using real and synthetic data sets.  \cite{bandaragoda2018isolation} point out that the ISO has weaknesses, including finding anomalies in multimodal data with a large number of modes and finding outliers that exist between axis-parallel clusters. The Isolation Forest algorithm uses decision trees that create random splits, which are either straight lines or hyperplanes perpendicular to one of the axes. Because these splits are axis-parallel, the algorithm may struggle to effectively separate outliers located between two clusters of data, unless the outlier happens to fall within the path of a split. \cite{bandaragoda2018isolation} address these weakness with a supervised method which we do not consider in this work. \cite{tan2022sparse} modify ISO to help with the identification of axis parallel outliers by first transforming data using random projections. } 

{Proximity-based or distance-based methods find outliers either based on a single distance measure, such as Mahalanobis distance, from some estimated center or using a local neighborhood of distances.  Local Outlier Factor (LOF) \citep{breunig2000lof} is a popular distance-based outlier detection algorithm that identifies outliers on the basis of local point densities. The implementation of LOF requires the specification of $k$, the number of neighbors.  Another widely implemented distance-based method is k-nearest neighbors (kNN) \citep{ramaswamy2000efficient} which ranks each point on the basis of its distance to its $k^{th}$ nearest neighbor.  From this ranking, one can declare the top $n$ points to be outliers. The kNN algorithm is  scalable and easy to understand but also requires the specification of $k$.  }

{Graph-based Learnable Unified Neighbourhood-based Anomaly Ranking (LUNAR) is a graph-neural network based outlier detection method that is more flexible and adaptive to a given set of data than local outlier methods such as kNN and LOF \citep{goodge2022lunar}. Different from the proximity based methods kNN and LOF, LUNAR learns to optimize model parameters for better outlier detection. LUNAR is more robust to the choice of $k$ compared to other local outlier methods and shows good performance when compared to other deep methods \citep{goodge2022lunar}. } 

{Deep learning methods such as Autoencoders or adversarial neural networks are increasingly used for unsupervised outlier detection \citep{chen2017outlier, lyudchik2016outlier, zhou2017anomaly, liu2019generative}. Deep-learning methods help to overcome the curse of dimensionality by learning the features of the data while simultaneously flagging anomalous observations. A competitive, deep-learning, unsupervised outlier detection method is Deep SVDD \citep{ruff2018deep}, which combines kernel-based SVDD  with deep learning to simultaneously learn the network parameters and minimize the volume of a data-enclosing hypersphere. Outliers are identified as observations whose distance are far from the estimated center. } 

{Deep autoencoders \citep{hinton2006reducing} are the predominant approach used in deep outlier detection.  Autoencoders are typically trained to minimize the reconstruction error, the difference between the input data and the reconstruction of the input data with the latent variable. A Variational Autoencoder (VAE) \citep{kingma2013auto} is a stochastic generative model in which the encoder and decoder are generated by probabilistic functions. In this manner VAE does not represent the latent space with a simple value but maps input data to a stochastic variable. Observations with a high reconstruction error are candidates to be considered anomalous in VAEs \citep{zhao2019pyod}. VAEs can suffer from mode collapse, where the model only generates a subset of the actual classes (one or two in severe cases).  This issue can lead to poor generalization and a lack of diversity in the generated data, which may cause the model to miss certain types of anomalies \citep{kossale2022mode}.}

{Probabilistic methods for unsupervised outlier detection estimate the density function of a data set.  One such method, Empirical Cumulative Distribution Functions (ECOD), \citep{li2022ecod} estimates the empirical distribution function of the data.  ECOD performs this estimation by computing an empirical cumulative distribution function for each dimension separately. \cite{li2022ecod} show ECOD to be a competitive method for outlier detection.  ECOD does not require hyperparameter specification and is a scalable algorithm. ECOD is based on non-parametric estimation, which can limit its flexibility in modeling complex distributions. If the data has complex structures or is multimodal, ECOD might not capture these complexities effectively, potentially missing anomalies \citep{li2022ecod}.}

{The sampling of outlier detection methods mentioned above, and others, from a variety of different approaches are implemented in the PyOD package \citep{zhao2019pyod}. Details of each implementation and the methods available can be found here: \url{https://github.com/yzhao062/pyod}. We have used the PyOD implementations of the above-mentioned algorithms for ease of comparison.}

{This paper is organized as follows Section \ref{sec:Boundary_Peeling} describes the Boundary Peeling One Class (BP) and the Ensembled version (EBP).  Section~\ref{sec:Example_data} compares the performance of Boundary Peeling with other benchmark methods on semantically constructed benchmark datasets.  Section~\ref{sec:syn_data} compares the performance of our method with other benchmark methods using synthetic data. Section~\ref{sec:discussion} discusses the limitations, contributions and future research. }

\section{Boundary Peeling}\label{sec:Boundary_Peeling}

Similar to the OCP method \citep{Martinez2020OCP} the Boundary Peeling One Class method paper uses successively peeled boundaries created one-class support vector machines (OCVM) \citep{scholkopf2001estimating}.  Interestingly, \cite{tax2004support} and \cite{ bounsiar2014one} show that One Class Support Vector Machines (OCSVM) are equivalent as long as the data is transformed to unit variance.  The OCP method uses these successively peeled boundaries to estimate a mean, which is then used to calculate distances. However, as shown by \cite{lee2022robustness}, this mean estimate can be biased, and the performance of the OCP method is heavily influenced by the underlying data distribution. Additionally, the OCP method struggles with handling datasets that contain more than one mode. The BP method addresses and corrects all of these limitations.





{Instead of employing a two-step process of mean estimation followed by distance calculation, the BP method calculates the signed distance from each observation to each successive OCSVM separating hyperplane. OCSVM uses optimization to determine the distance from a hyperplane to the origin. Using a  kernel function, $K(\xvec_i, \xvec_j)$, typically a Gaussian kernel function for OCSVM, the separating boundary can be found by solving for the Lagrange multipliers $\alpha_i$.}

\begin{equation}
\label{eq:ocsvm}
\begin{aligned}
\underset{\alphavec}{\text{minimize}} \quad & \frac{1}{2} \sum_{i,j} \alpha_i \alpha_j K(\xvec_i, \xvec_j)\\
\text{subject to} \quad & 0 \leq \alpha_i \leq \frac{1}{\nu n}, \quad \sum_i \alpha_i =1, \quad \forall i=1,\ldots,n
\end{aligned}
\end{equation}

{\noindent Equation \ref{eq:ocsvm} gives the optimization formulation after applying Lagrangian multipliers. Here $\nu$ is a parameter that controls the trade off between the number of examples in the training set identified as outliers and $n$ is the sample size.  Once the optimization above is solved and a set of weights, $\alpha_i$ are obtained then the decision function for any new observation $\xvec_*$ is }

\begin{align}
    f(\xvec_*) = \text{sgn}(\sum_i \alpha_i K(\xvec_i, \xvec_*)-\rho)
\end{align}

\noindent { where $\rho$ is the distance from the origin to the hyperplane.  A positive value of the decision function indicates that a sample is part of the training set or an inlier, while a negative value signifies that the test point is an outlier. If $f(\mathbf{x}_i) = 0$, the observation is considered a support vector. For more details on OCSVM, refer to \cite{scholkopf2001estimating}. In our implementation of OCSVM, we use the Gaussian kernel function and set the bandwidth parameter, $s = p$, as recommended in \cite{lee2022robustness} and \cite{tax2004support}.}




{In the implementation of the BP method, each observation receives a decision function value from every successively created boundary, denoted as $\hat{f(x)_{ij}}$, even if the observation has been removed from a previous boundary. The BP method continues to create boundaries until only two observations remain, with $j = 1, \ldots, \textit{peel}$ representing the boundaries. The decision function values, resulting in an $n \times \textit{peel}$ matrix, are then averaged for each observation $i$, yielding $\overline{f(x)}_i$.}

{Observations with values of $\hat{f(x)_{i}} \geq 0$ in the early (or outermost) peels are more likely to be outliers and are therefore inversely weighted. For example, if observation $i$ is labeled as an outlier (support vector) during the first peel, then it would have a depth of $depth_i = 1$. If a total of 10 peels are constructed to eliminate all but 2 observations, then $peel = 10$. The weight $d$ for this observation would be $d_i = \frac{peel}{depth_i} = \frac{10}{1} = 10$, a relatively large weight since an early peel indicates a potentially anomalous observation. The weighted average of the decision function values for each observation results in a vector of final kernel distance scores of length $n$, given by $\text{KDS}_i = d_i \cdot \overline{f(x)}_i$. An observation is flagged as an outlier if its $\text{KDS}_i$ exceeds a threshold defined as $h = Q_3(\mathbf{KDS}) + 1.5 \cdot \text{IQR}(\mathbf{KDS})$, where $Q_3$ is the third quartile and $\text{IQR}$ is the interquartile range. All steps are detailed in Algorithm \ref{alg:BP}.}

\begin{algorithm}[H] \label{alg:BP}
\renewcommand{\thealgorithm}{}
\caption{BP}

\begin{algorithmic}[1]
\Procedure{BP}{$S, q = 0.01, n_{peel} = 2$}
\State $r\gets n$
\State $s\gets p$
\State $S_{r}\gets S$
\State $peel\gets 0$
\While{$r > n_{peel}$} 
\State $X_{SV} \gets \text{OCSVM}(S_{r},q,s)$
\State $S_{r}\gets S_{r}\symbol{92}X_{SV}$
\State $D_r \gets f(S)$
\State $r\gets$ rows$(S_{r})$
\State $peel\gets$ $peel + 1$
\State $depth_i\gets$ $peel$
\EndWhile\label{euclidendwhile}
\State $\text{d} \gets$ $ peel / depth_i$
\State $\text{KDS}_i \gets d \cdot mean(D_r)$
\State $h \gets  Q_3\mathbf{(KDS)}+1.5\cdot \text{IQR}\mathbf{(KDS)}$ 
\State $\text{flag}_{\text{BP}} \gets I(\text{KDS}_i > h)$ 

\State \textbf{return} $\text{KDS}_i,\text{flag}_{\text{BP}}$
\EndProcedure
\end{algorithmic}
\end{algorithm}

\begin{figure*}[!t]
    \centering
    \includegraphics[width=165mm]{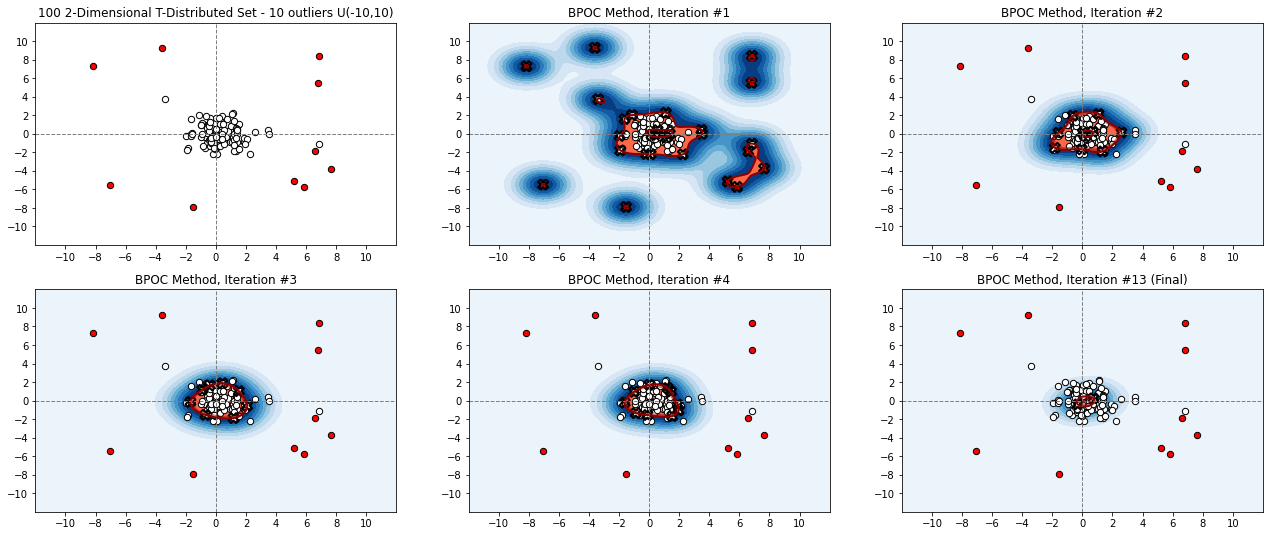}
    \caption{BP Example on a 2-dimensional unimodal data set. There are 100 inlier observations generated by $t(df = 5)$, and 10 outlier observations generated from $U(-10,10)$. Contours indicate kernel signed distances from separating hyperplane. Blue contours indicate positive distances (darker blue indicates distance closer to zero), while red indicate negative distances. Support vectors are marked with an X.}
    \label{fig:2dsimT_ex}
\end{figure*}

{To illustrate how the BP method works, we plotted the boundaries generated from peels on 100 observations sampled from a multivariate t-distribution (with $df=5$) as inliers and 10 observations from a uniform distribution $U(-10, 10)$ as outliers (see Figure \ref{fig:2dsimT_ex}).} The second image in the top row shows the first flexible boundary. Each outlier is indicated by a blue contour, representing a positive distance, meaning those observations lie outside the hyperplane. In the second iteration (Figure~\ref{fig:2dsimT_ex}, third image), the outliers and some inliers from the outer regions across all four quadrants no longer have contours around them. These observations were identified as support vectors and were "peeled" off before constructing the second boundary. The blue contours indicating a positive distance are now more concentrated in the center of the distribution. In the bottom row of Figure~\ref{fig:2dsimT_ex}, the left and center images show that peels 3 and 4 each remove only a small number of points. The final image on the far right of the bottom row depicts the last peel, where only a few points remain. At this stage, the 13 signed distance values from each peel are weighted and averaged for each observation.

\begin{figure*}[!t]
    \centering
    \includegraphics[width=165mm]{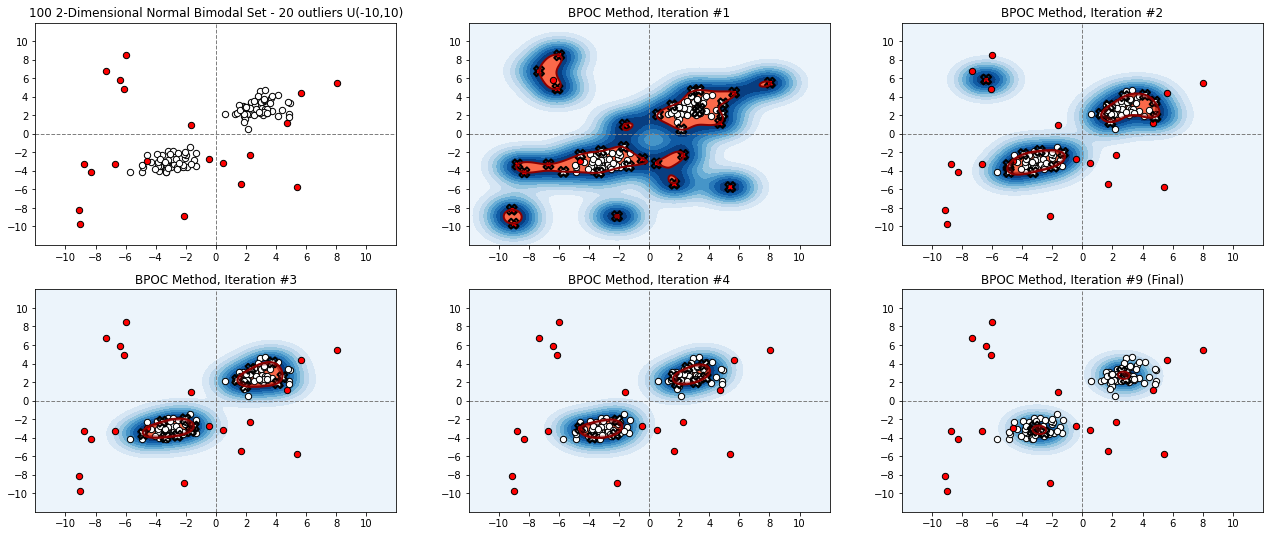}
    \caption{BP Example on a 2-dimensional bimodal data set. 50 inlier observations generated from $N(-3,1)$ and 50 inlier observations generated from $N(3,1)$. 20 outlier observations generated from $U(-10,10)$. Contours indicate kernel signed distances from separating hyperplane. Blue contours indicate positive distances (darker blue indicates distance closer to zero), while red indicate negative distances. Support vectors are marked with an X.}
    \label{fig:2dsimN_ex}
\end{figure*}

Similarly, Figure \ref{fig:2dsimN_ex} demonstrates the BP method's application to a bimodal dataset. Fifty inlier observations are generated from a normal distribution with a mean vector $\boldsymbol{\mu} = \mathbf{-3}$ and off-diagonal elements of the covariance matrix $\Sigma$ set to 0.5. For the second mode, another 50 observations are generated from a normal distribution with a mean vector $\boldsymbol{\mu} = \mathbf{3}$ and off-diagonal elements of $\Sigma$ also equal to 0.5. The 20 outlier observations are sampled from a uniform distribution $U(-10, 10)$. In the first two iterations, we observe that outlying observations are identified as support vectors. As the process continues, inlier observations, which have lower kernel distances, are peeled last, reflecting their proximity to the modes. Consequently, the average kernel distances for the outliers are higher, as they are located farther from the modes.

\subsection{Ensembled Boundary Peeling}

{To increase the sensitivity of the BP method, we computed the average distance after applying the BP method to datasets with randomly sampled columns, denoted as $Y_r$.} This approach is referred to as the ensemble BP (EBP). While the BP algorithm (Algorithm~\ref{alg:BP}) is relatively fast compared to other methods (see Table \ref{tab:time}), the EBP algorithm, though slower, remains a strong-performing and feasible option. The primary difference in the EBP algorithm lies in line 5 of Algorithm \ref{alg:ensemble_ocp}, where a subset of $\sqrt{p}$ features is selected, and the algorithm is repeated $c$ times. Feature bagging has been shown to enhance outlier detection performance by decorrelating predictions \citep{lazarevic2005feature}. In our implementation, we set $c=50$ to balance computational efficiency during simulations. The EBP flags outliers by using the average $\text{KDS}_i = d_i \cdot \overline{f(x)}_i$ from each of the $c$ ensembles, compared against a robust threshold $h$, calculated from the $\text{KDS}_i$ values averaged over all $c$ iterations.

\begin{algorithm}[!h]\label{alg:ensemble_ocp}
\renewcommand{\thealgorithm}{}
\caption{The Ensemble BP Method}
\begin{algorithmic}[1]
\Procedure{EBP}{$S,q = 0.01, n_{peel} = 2,c=50$}
\State $r\gets n$
\State $S_{r}\gets S$
\For{$i \gets 1$ to $c$} 
\State $Y_{r}\gets \text{colsample}(S_{r}, int(\sqrt{p}))$
\While{$r > n_{peel}$}
\State $X_{SV} \gets \text{OCSVM}(Y_{r},q,s)$
\State $Y_{r}\gets Y_{r}\symbol{92}X_{SV}$
\State $D_r \gets f(S)$
\State $r\gets \text{rows}(Y_{r})$
\State $peel\gets$ $peel + 1$
\State $depth_i\gets$ $peel$
\EndWhile\label{euclidendwhile}
\State $\text{d} \gets$ $depth_i / peel$
\State $\text{KDS}_i \gets \text{mean}(D_{r})$
\EndFor
\State $\text{EKDS}_i \gets \text{mean}(\text{KDS}_i) \times \text{mean}(d)$ 
\State $h=Q_3\mathbf{(\text{EKDS})}+1.5\cdot \text{IQR}\mathbf{(\text{EKDS})}$  
\State $\text{flag}_{\text{EBP}} \gets I(\text{ERKD}_i > h)$ 
\State \textbf{return} $\text{ERKD}_i,\text{flag}_{\text{EBP}}$
\EndProcedure
\end{algorithmic}
\end{algorithm}

\section{Synthetic Data Comparison}\label{sec:syn_data}

To investigate the behavior of the BP and EBP methods under controlled conditions, we conducted simulations on 1,000 datasets, each with a sample size of $n = 50$ and a dimensionality of $p = 100$. Observations were drawn from various distributions, including multivariate normal, t-distribution (with $df = 5$), lognormal, and Wishart distributions, and were applied to both unimodal and multimodal data. For each of the 1,000 iterations, sample data were generated randomly with different correlation structures—none, medium, and high—for scenarios both without outliers and with 10\% outliers. In the case of unimodal data with a single mode, the data were generated with a $p$-dimensional mean vector $\boldsymbol{\mu} = \mathbf{0}$. For bimodal data, the second mode was generated with a mean vector $\boldsymbol{\mu} = \mathbf{5}$. When no correlation was present, the covariance matrix $\Sigma$ was set to the identity matrix $I$. For moderate and high correlation, the off-diagonal elements of $\Sigma$ were set to 0.5 or 0.75, respectively. Outliers for the normal, t, and Wishart distributions were uniformly generated from $U(-10, 10)$, while for the lognormal distribution, outliers were generated from $U(-20, 20)$. Additionally, a mixed distribution scenario was included, where both the correlation and the distributions for each mode were chosen randomly, with correlation values of 0, 0.5, or 0.75, and distributions selected from the aforementioned types.

For a more realistic scenario, we also conducted a simulation where the number of modes, the distribution of each mode, the means and covariances of the modes, and the percentage of outliers were randomly generated for each iteration. In each iteration, the sample size $n$ was randomly chosen from the range $[50, 150]$ and the dimensionality $p$ from the range $[50, 300]$. The number of modes was randomly selected between 1 and 5, with $n$ evenly distributed among the modes. The off-diagonal elements of the covariance matrix $\Sigma$ were uniformly chosen between 0 and 1. Data for each mode was randomly generated from one of the following distributions: multivariate normal, t-distribution (with $df = 5$), lognormal, or Wishart. The percentage of contamination was randomly assigned to one of three levels: no contamination, 1\% to 10\%, or 10\% to 20\%. Outliers were generated from a uniform distribution $U(-20, 20)$.

For all scenarios, we compared BP and EBP methods with several other outlier detection algorithms, including {OCP}, ISO, ECOD, kNN, LOF, LUN, VAE, and DSVDD. {All methods except OCP} were implemented using the default or recommended settings provided in the PyOD package \cite{zhao2019pyod}.  {The OCP method was implemented using the robust threshold as described in \cite{Martinez2020OCP}.} The parameters for BP and EBP were set as outlined in Algorithms \ref{alg:BP} and \ref{alg:ensemble_ocp}. The contamination ratio ($cr$) was consistently maintained at a default value of 10\% for all competing algorithms.

We measured performance using several metrics: detection rate (DR), correct classification rate (CC), area under the curve (AUC), and precision (PREC). Using the standard definitions for True Positive (TP), True Negative (TN), False Positive (FP), and False Negative (FN), we define the detection rate for dataset $i$ as $DR_i = \frac{TP_i}{TP_i + FN_i} \times 100$. The correct classification rate for a dataset $i$ is defined as $\text{CC}_i = \frac{TP_i + TN_i}{n} \times 100$, while precision is defined as $\text{PREC}_i = \frac{TP_i}{TP_i + FP_i} \times 100$. The Area Under the Curve (AUC) of the Receiver Operating Characteristic (ROC) curve measures the probability of correctly distinguishing inliers from outliers. For brevity, only the tables for CC and AUC are presented here; tables for DR and PREC are available in the supplementary materials.

\begin{table}[H]
\small
\caption{\label{tab:0_percent_out}Correct Classification rates for unimodal and bimodal simulated data with no outliers for $n=50$ and $d=100$. The parenthesis indicate the mean(s) of the mode ($\boldsymbol\mu_1$) or modes ($\boldsymbol\mu_1,\boldsymbol\mu_2$). Bold indicates the best
performance. {N indicates the data was sampled from a multivariate normal distribution.  LN indicates the data was sampled from a multivariate log normal distribution.  t indicates the data was sampled from a multivariate t distribution with $df$=5.  W indicates the data are sampled from a multivariate Wishart distribution. }}

\centering
\begin{tabular}{llllllllllll} \toprule
             & Cor & BP   & EBP & {OCP} & ISO & ECOD & LOF & kNN    & DSVDD & LUNAR & VAE \\ \cmidrule(r){1-12}
             
N(\textbf{0})    & 0  &\textbf{100.00} &	94.856 &\textbf{100.00}  &	99.824	& 90.000	&90.000	& 90.002 &	90.000& 	90.000 &	90.000
  \\
LN(\textbf{0})    & 0 & 98.284	& 95.890	& \textbf{100.00} & 99.996    & 90.000   & 90.000   & 90.000      & 90.000         & 90.000     & 90.000   \\
t(\textbf{0})     & 0 & 99.948 & 99.539   & \textbf{100.00}  & 90.339     & 90.000    & 90.000   & 90.000      & 90.000         & 90.000     & 90.000   \\
W(\textbf{0})     & 0 & \textbf{100.00}    & 98.730   &  \textbf{100.00}   & 99.794     & 90.000    & 90.000   & 90.000      & 90.000         & 90.000     & 90.000  \\ \bottomrule
N(\textbf{0})    & 0.5  & 94.928 &	84.242	& \textbf{99.254} & 89.950 &	90.000 &	90.000&	90.010&	90.000&	90.000 &	90.000
   \\
LN(\textbf{0})    & 0.5 &\textbf{100.00} &	83.720 & 99.952 &	88.864   &90.000    & 90.000   & 90.000      & 90.000         & 90.000     & 90.000   \\
t(\textbf{0})     & 0.5 & 99.440 &    95.979  & \textbf{100.00} &   88.238   & 90.000    & 90.000   & 90.003      & 90.000         & 90.000     & 90.000   \\
W(\textbf{0})     & 0.5 & \textbf{100.00}    &  98.785    &  \textbf{100.00} & 99.800    & 90.000    & 90.000   & 90.000      & 90.000         & 90.000     & 90.000  \\ \bottomrule
N(\textbf{0})    & 0.75  &  88.778 &	83.932 & \textbf{92.969}	& 86.430  & 90.000   & 90.000   & 90.034 & 90.000         & 90.000     & 90.000   \\
LN(\textbf{0})    & 0.75 & 82.641 &  90.742  & \textbf{99.876} & 87.104   &90.000    & 90.000   & 90.000      & 90.000         & 90.000     & 90.000   \\
T(\textbf{0})     & 0.75 &  92.414    & 90.830  & \textbf{99.759}  & 86.790 & 90.000    & 90.000   & 90.009      & 90.000         & 90.000     & 90.000   \\
W(\textbf{0})     & 0.75 &  \textbf{100.00}  &  98.788  & \textbf{100.00} &  99.803  & 90.000    & 90.000   & 90.001     & 90.000         & 90.000     & 90.000  \\ \midrule

N(\textbf{0,5})    & 0  & 99.994	& 96.981 & \textbf{100.00} &	98.433 &	90.000 & 90.000 & 90.740  & 90.000  &90.000         & 90.000    \\
LN(\textbf{0,5})    & 0 & \textbf{99.912} & 	96.742 & 95.952	& 99.942  &90.000    & 90.000   & 90.000      & 90.000         & 90.000     & 90.000   \\
t(\textbf{0,5})     & 0 &   \textbf{100.00} &	86.444 &	94.156& 88.492  & 90.000    & 90.000   & 90.009      & 90.000         & 90.000     & 90.000   \\
W(\textbf{0,5})     & 0 & \textbf{100.00} &	98.022 & 99.973	& 99.686  & 90.000    & 90.000   & 90.006     & 90.000         & 90.000     & 90.000  \\ \bottomrule
N(\textbf{0,5})    & 0.5  &  \textbf{99.238}	& 87.376 &93.818 & 86.536 & 90.000   & 90.000   & 90.002 & 90.000         & 90.000     & 90.000   \\
LN(\textbf{0,5})    & 0.5 & \textbf{100.00}	& 86.880 & 86.230&	87.834   &90.000    & 90.000   & 90.001      & 90.000         & 90.000     & 90.000   \\
t(\textbf{0,5})     & 0.5 & \textbf{100.00} &	86.146 & 92.885&	85.966 & 90.000    & 90.000   & 90.003      & 90.000         & 90.000     & 90.000   \\
W(\textbf{0,5})     & 0.5 &  \textbf{100.00} &	98.060 &  \textbf{100.00}&	99.696 & 90.000    & 90.000   & 90.001     & 90.000         & 90.000     & 90.000  \\ \midrule
N(\textbf{0,5})    & 0.75  & \textbf{94.954}	& 86.470 &  93.309 & 82.316 & 90.000   & 90.000   & 90.033 & 90.000         & 90.000     & 90.000   \\
LN(\textbf{0,5})    & 0.75 &  \textbf{100.00} &	84.856 &84.078 &	85.190   &90.000    & 90.000   & 90.000      & 90.000         & 90.000     & 90.000   \\
t(\textbf{0,5})     & 0.75 &  \textbf{99.752} &	86.108 & 91.764&	84.618 & 90.000    & 90.000   & 90.004      & 90.000         & 90.000     & 90.000   \\
W(\textbf{0,5})     & 0.75 &  \textbf{100.00} &	98.120 &99.998&	99.712  & 90.000    & 90.000   & 90.000     & 90.000         & 90.000     & 90.000  \\ \midrule
Average & & \textbf{97.928} &	92.010 &	96.832 &	92.306 &	90.000	 &90.000	&90.036	&90.000&	90.000&	90.000\\ \midrule
{Rank}  & & 1 &	4	&2	&3&	9&	9&	5&	9&	9&	9\\ \bottomrule

\end{tabular}

\end{table}

The results in Table~\ref{tab:0_percent_out} demonstrate that when no outliers are present, BP achieves the best and most consistent correct classification rate {and improves upon the OCP method}. This holds true for both unimodal and bimodal datasets with two modes (where $\boldsymbol{\mu}_1 = 0$ and $\boldsymbol{\mu}_2 = 5$). For smaller sample sizes, this property will help to protect against unnecessary data loss. {The OCP method  preforms well when no outliers are present, however the performance of the OCP method degrades with the increasing correlation.  We should note that most likely the OCP method is not recognizing both modes, but simply placing the mean estimate in between the modes. Table~\ref{tab:10_percent_out_AUC} below verifies this.}  It is also important to note that for many methods, the PyOD package requirement to specify a percentage of outliers in advance leads to the incorrect classification of 10\% of points as outliers, regardless of the actual data distribution.

\begin{table}[H]
\small
\caption{\label{tab:10_percent_out_CC}Correct Classification rates for unimodal and bimodal simulated data with 10\% outliers for $n=50$ and $d=100$. The parenthesis indicate the mean(s) of the mode ($\boldsymbol\mu_1$) or modes ($\boldsymbol\mu_1,\boldsymbol\mu_2$). Bold indicates the best
performance. { N indicates the data was sampled from a multivariate normal distribution.  LN indicates the data was sampled from a multivariate log normal distribution.  t indicates the data was sampled from a multivariate t distribution with $df$=5.  W indicates the data are sampled from a multivariate Wishart distribution.}}

\begin{tabular}{@{}p{1.5cm}llllllllllll@{}}  \toprule
        & Cor  & BP     & EBP  & {OCP}   & ISO     & ECOD   & LOF    & kNN    & DSVDD  & LUN    & VAE    \\ \midrule
N(\textbf{0})    & 0    & \textbf{100.00} & 99.793 & 98.546 & \textbf{100.00} & 93.333 & 88.367 & 93.167 & 79.440 & 93.333 & 93.333 \\
N(\textbf{0},\textbf{5})  & 0    & 99.111 & 98.909 &99.715 & \textbf{100.00} & 99.178 & 99.262 & 99.476 & 83.455 & 98.953 & 98.965 \\
LN(\textbf{0})   & 0    & 99.357 & 99.887 &90.938 & \textbf{100.00} & 93.333 & 88.133 & 93.160 & 79.360 & 93.333 & 93.333 \\
LN(\textbf{0},\textbf{5}) & 0    & 98.348 & 99.922 &97.334 & \textbf{100.00} & 88.580 & 87.660 & 93.120 & 76.843 & 93.327 & 93.475 \\
t(\textbf{0})    & 0    & 99.263 & 97.837 & 90.909 & \textbf{99.757} & 93.320 & 88.253 & 93.147 & 79.200 & 93.333 & 93.333 \\
t(\textbf{0},\textbf{5})  & 0    & 98.338 & 98.538 & 96.398 & \textbf{99.75}2 & 89.478 & 88.250 & 93.103 & 77.167 & 93.327 & 93.285 \\
W(\textbf{0})    & 0    & 99.867 & 99.867 & 92.324 & \textbf{100.00} & 93.333 & 88.700 & 92.967 & 76.600 & 93.333 & 93.333 \\
W(\textbf{0},\textbf{5})  & 0    & 99.498 & 99.995 &99.962 & \textbf{100.00} & 89.482 & 88.300 & 92.862 & 76.590 & 93.333 & 93.347 \\ \midrule
N(\textbf{0})    & 0.5  & 99.757 & 97.717 & 96.088  & \textbf{99.993} & 93.320 & 87.880 & 92.863 & 80.033 & 93.333 & 93.333 \\
N(\textbf{0},\textbf{5})  & 0.5  & 95.180 & 93.085 & 97.080 & \textbf{99.885} & 98.991 & 99.244 & 99.385 & 83.938 & 99.065 & 99.073 \\
LN(\textbf{0})   & 0.5  & 98.100 & 86.653 &90.909& \textbf{99.647} & 93.333 & 86.463 & 92.717 & 80.887 & 93.327 & 93.327 \\
LN(\textbf{0},\textbf{5}) & 0.5  & 97.658 & 84.512 &94.045 & \textbf{99.825} & 88.493 & 89.813 & 93.193 & 78.273 & 93.297 & 93.457 \\
t(\textbf{0})    & 0.5  & 99.237 & 97.393 & 90.909 & \textbf{99.430} & 93.093 & 87.373 & 92.517 & 80.067 & 93.307 & 93.300 \\
t(\textbf{0},\textbf{5})  & 0.5  & 98.610 & 97.940 &96.723 & \textbf{99.475} & 86.848 & 88.107 & 92.633 & 77.177 & 93.333 & 93.062 \\
W(\textbf{0})    & 0.5  & 99.800 & \textbf{100.00} & 92.087& \textbf{100.00} & 93.333 & 88.767 & 92.900 & 76.667 & 93.333 & 93.333 \\
W(\textbf{0},\textbf{5})  & 0.5  & \textbf{100.00} & \textbf{100.00} & 99.975& \textbf{100.00} & 88.767 & 88.367 & 92.767 & 76.400 & 93.333 & 93.533 \\ \midrule
N(\textbf{0})    & 0.75 & 99.867 & 97.517 & 95.345 & \textbf{99.897} & 93.167 & 87.057 & 92.707 & 84.093 & 93.333 & 93.333 \\
N(\textbf{0},\textbf{5})  & 0.75 & 95.751 & 91.698 & 97.883 & \textbf{99.425} & 98.442 & 99.253 & 99.316 & 84.585 & 99.073 & 99.024 \\
LN(\textbf{0})   & 0.75 & 97.720 & 86.343 &90.909 & \textbf{99.147} & 93.300 & 85.030 & 92.440 & 83.153 & 93.287 & 93.273 \\
LN(\textbf{0},\textbf{5}) & 0.75 & 97.190 & 90.415 &94.785 & \textbf{99.503} & 88.428 & 90.532 & 93.193 & 79.213 & 93.253 & 93.417 \\
t(\textbf{0})    & 0.75 & 99.257 & 97.110 & 92.254& \textbf{99.100} & 92.520 & 86.930 & 92.503 & 83.033 & 93.333 & 93.333 \\
t(\textbf{0},\textbf{5})  & 0.75 & 98.478 & 97.688 &96.585  &\textbf{99.158} & 84.060 & 88.665 & 92.522 & 78.083 & 93.317 & 92.828 \\
W(\textbf{0})    & 0.75 & 99.867 & 99.933 & 92.283 & \textbf{100.00} & 93.333 & 88.867 & 92.900 & 77.200 & 93.333 & 93.333 \\
W(\textbf{0},\textbf{5})  & 0.75 & \textbf{100.00} & \textbf{100.00} & 99.980& \textbf{100.00} & 88.867 & 88.767 & 92.833 & 76.867 & 93.333 & 93.500 \\ \midrule
\textbf{Mixed}(\textbf{0},\textbf{5})  &      & 93.920 & \textbf{98.243} & 95.656& 94.297 & 87.810 & 88.380 & 89.703 & 78.227 & 89.873 & 89.967 \\ \midrule
\textbf{Average} &      & 98.567& 96.440 & 95.185 & \textbf{99.532} & 91.846 &89.457 &93.524 & 79.462 & 93.869 & 93.873 \\ \midrule
{Rank} & & 2	 &3&	4&	1&	8&	9&	7&	10&	6&	5 \\

\bottomrule
\end{tabular}

\end{table}

\begin{table}[H]
\small
\caption{\label{tab:10_percent_out_AUC}AUC for unimodal and bimodal simulated data with no outliers for $n=50$ and $d=100$. The parenthesis indicate the mean(s) of the mode ($\boldsymbol\mu_1$) or modes ($\boldsymbol\mu_1,\boldsymbol\mu_2$). Bold indicates the best
performance. {N indicates the data was sampled from a multivariate normal distribution.  LN indicates the data was sampled from a multivariate log normal distribution.  t indicates the data was sampled from a multivariate t distribution with $df$=5.  W indicates the data are sampled from a multivariate Wishart distribution. }}
\begin{tabular}{@{}p{1.5cm}llllllllllll@{}} \toprule
        & Cor  & BP    & EBP   & {OCP}   & ISO     & ECOD    & LOF     & kNN     & DSVDD  & LUN    & VAE     \\  \midrule
N(\textbf{0})    & 0    & \textbf{100.00 }& \textbf{100.00} & 99.843  & \textbf{100.00} & \textbf{100.00} & \textbf{100.00} & \textbf{100.00} & 46.798 & 99.736 & 99.652 \\
N(\textbf{0},\textbf{5})  & 0    & \textbf{100.00} & \textbf{100.00} & 99.201  & \textbf{100.00} & \textbf{100.00} & \textbf{100.00} & 100.00 & 31.707 & 99.573 & \textbf{100.00} \\
LN(\textbf{0})   & 0    & \textbf{100.00} & \textbf{100.00} &98.534  & \textbf{100.00} & \textbf{100.00} & \textbf{100.00} & \textbf{100.00 }& 49.665 & 99.822 & 99.734 \\
LN(\textbf{0},\textbf{5}) & 0    & 99.984 & \textbf{100.00} & 50.259& \textbf{100.00} & \textbf{100.00} & 99.996 & 99.996 & 27.119 & 99.804 & \textbf{100.00} \\
t(\textbf{0})    & 0    & 99.998 & \textbf{99.999} & 98.019  & 99.994 & 99.992 & 99.998 & 99.998 & 51.463 & 99.676 & 99.668 \\
t(\textbf{0},\textbf{5})  & 0    & \textbf{99.989} & 99.950 & 50.000 & 99.950 & 99.986 & 99.988 & 99.988 & 36.929 & 99.700 & 99.945 \\
W(\textbf{0})    & 0    & \textbf{100.00} & \textbf{100.00 }&99.979   & \textbf{100.00} & \textbf{100.00} & \textbf{100.00} & \textbf{100.00} & 30.444 & 99.400 & 98.656 \\
W(\textbf{0},\textbf{5})  & 0    & \textbf{100.00} & \textbf{100.00} &57.902 & \textbf{100.00} & \textbf{100.00} & \textbf{100.00} & \textbf{100.00} & 30.347 & 99.036 & 99.981 \\ \midrule
N(\textbf{0})    & 0.5  & \textbf{100.00} & \textbf{100.00}& 98.394  & \textbf{100.00} & 99.994 & \textbf{100.00} & \textbf{100.00} & 47.930 & 99.978 & 99.979 \\
N(\textbf{0},\textbf{5})  & 0.5  & \textbf{100.00} & \textbf{100.00 }&97.849& \textbf{100.00} & 99.973 & \textbf{100.00} & \textbf{100.00} & 38.245 & 99.844 & \textbf{100.00} \\
LN(\textbf{0})   & 0.5  & 99.991 & 99.900 &77.676  & 99.999 & \textbf{100.00} & 99.996 & 99.991 & 63.800 & 99.881 & 99.811 \\
LN(\textbf{0},\textbf{5}) & 0.5  & 99.813 & 98.789 &50.000 & 98.789 & 99.961 & \textbf{99.998} & 99.975 & 43.632 & 99.827 & 99.978 \\
t(\textbf{0})    & 0.5  & \textbf{99.977} & 99.970 & 98.198 & 99.960 & 99.844 & 99.984 & \textbf{99.977} & 51.867 & 99.737 & 99.747 \\
t(\textbf{0},\textbf{5})  & 0.5  & 99.991 & 99.899 & 50.243& 99.899 & 97.589 & 99.998 & \textbf{100.00} & 45.483 & 99.869 & 99.820 \\
W(\textbf{0})    & 0.5  & \textbf{100.00} & \textbf{100.00} &99.986 & \textbf{100.00} & \textbf{100.00} & \textbf{100.00} & \textbf{100.00} & 29.668 & 98.548 & 99.024 \\
W(\textbf{0},\textbf{5})  & 0.5  & \textbf{100.00} & \textbf{100.00}& 57.174 & \textbf{100.00} & \textbf{100.00} & \textbf{100.00} & \textbf{100.00} & 26.775 & 99.363 & \textbf{100.00} \\ \midrule
N(\textbf{0})    & 0.75 & \textbf{100.00} & \textbf{100.00} &98.836& \textbf{100.00} & 99.892 & \textbf{100.00} & 100.00 & 63.888 & 99.990 & 99.987 \\
N(\textbf{0},\textbf{5})  & 0.75 & \textbf{100.00} & \textbf{100.00} &97.440 & \textbf{100.00} & 99.696 & \textbf{100.00} & \textbf{100.00} & 49.808 & 99.953 & \textbf{100.00} \\
LN(\textbf{0})   & 0.75 & 99.946 & 99.535 &90.597 & 99.969 & 99.979 & \textbf{99.987} & 99.943 & 73.388 & 99.872 & 99.853 \\
LN(\textbf{0},\textbf{5}) & 0.75 & 99.668 & 98.252  & 50.000 & 98.252 & 99.759 & \textbf{99.987} & 99.956 & 53.806 & 99.811 & 99.942 \\
t(\textbf{0})    & 0.75 & 99.998 & 99.992 & 98.269 & 99.977 & 99.504 & 99.998 & \textbf{99.999} & 64.051 & 99.908 & 99.956 \\
t(\textbf{0},\textbf{5})  & 0.75 & 99.990 & 99.859 &58.409 & 99.859 & 94.978 & 99.992 & \textbf{99.994} & 49.713 & 99.908 & 99.686 \\
W(\textbf{0})    & 0.75 & \textbf{100.00} & \textbf{100.00} &99.986 & \textbf{100.00} & \textbf{100.00} & \textbf{100.00} & \textbf{100.00} & 31.076 & 99.040 & 98.552 \\
W(\textbf{0},\textbf{5})  & 0.75 & \textbf{100.00} & \textbf{100.00} &  57.973 & \textbf{100.00} & \textbf{100.00} & \textbf{100.00} & \textbf{100.00} & 28.714 & 98.592 & \textbf{100.00} \\ \midrule
\textbf{Mixed}(\textbf{0},\textbf{5})   &      & 99.880 & 99.884 & 99.924 & 99.884 & 98.818 & \textbf{99.944} & 99.920 & 67.938 & 99.806 & 99.939 \\ \midrule
Average &      & 99.969	&99.841	& 81.388	&99.861&	99.599	&\textbf{99.995}	&99.989	&45.370&	99.627	&99.756\\
\midrule
{Rank} & & 3	 &5&	9&	4&	8&	1&	2&	10&	7&	6 \\
 \bottomrule
\end{tabular}

\end{table}

. 

Table~\ref{tab:10_percent_out_CC} presents the correct classification (CC) rates for unimodal data with $\boldsymbol{\mu} = 0$ and bimodal data with $\boldsymbol{\mu}_1 = 0$ and $\boldsymbol{\mu}_2 = 5$. ISO achieves the highest average CC, closely followed by BP, which performs particularly well for data with low to moderate correlation. For data with moderate correlation, BP, EBP, and ISO all achieve perfect CC in the bimodal Wishart distribution, with EBP and ISO also attaining perfect CC in the unimodal case. LOF delivers the second-highest CC for bimodal, moderately correlated normal data. When the correlation is high, BP and EBP methods perform best with non-normal data. In contrast, the OCP method shows average performance but struggles with skewed lognormal data. Overall, there is no significant difference in CC between unimodal and bimodal data. {However, EBP exhibits slightly lower CC and detection rate (DR) for highly correlated lognormal data, suggesting that EBP may fail to identify many outliers in skewed and highly correlated datasets. This issue could potentially be addressed by adjusting the outlier threshold, though further investigation is needed.} When the correlation and distributions of the two modes are mixed, EBP achieves the highest CC rate. In summary, ISO leads with the highest average CC, followed by BP and EBP, while DSVDD shows the lowest average CC.

Table~\ref{tab:10_percent_out_AUC} shows that all methods, except for DSVDD {and OCP}, achieve high AUC values, indicating that at some threshold, each method can almost perfectly identify outliers in this simulated data. LOF has the highest overall average AUC, followed by kNN, BP, and VAE. ISO and EBP perform equally well in the mixed distribution case, with BP closely trailing. {It is worth noting that the OCP method shows AUC values around 50 in all bimodal cases, suggesting that the algorithm is mistakenly identifying one of the modes as outliers.}

Although not included in the main body of the paper, the detection rate (DR) is high for most methods across many scenarios {except for OCP} (see Supplementary Materials). Data with two normally distributed modes, regardless of the correlation structure, yields the highest DR. ISO achieves a perfect DR in all cases except for those involving mixed distributions. {OCP struggles particularly with heavy-tailed and skewed data, as indicated by several instances of DR=0.} {In every scenario where ISO achieves a DR of 100\%, BP or EBP also achieve a DR of 100\% or at least 99.272\%.} EBP has the highest DR when the modes are randomly distributed. ISO has the highest overall average DR at 99.737, closely followed by BP at 99.724, while OCP shows the lowest DR overall. ISO also has the highest average precision, followed by kNN, LUN, and VAE (see Supplementary Materials). {The top eight methods all have a precision greater than 98\%.} Interestingly, several methods, including BP and EBP, achieve perfect precision when the modes are generated from the Wishart distribution with high correlation. {Generally, a method with lower precision but a high detection rate tends to classify more inliers as outliers; this behavior can potentially be adjusted by modifying the threshold, where applicable.} 



Table~\ref{tab:random_sim} presents the results of the random simulation {for all methods except for OCP. OCP was not included given the poor AUC and DR on bimodal and skewed data.} When no outliers are present, BP achieves the highest correct classification (CC) rate, followed by EBP. This pattern also holds when the percentage of outliers is between 1\% and 10\%. At higher percentages of outliers, ISO achieves the best CC rate, followed by LOF. For datasets containing 1-10\% outliers, ISO, LOF, kNN, LUN, and VAE all achieve a detection rate (DR) of 100.00\%. However, at higher percentages of outliers, only kNN maintains a DR of 100.00\%. BP and EBP exhibit mid-level DRs of 98.208\% and 98.734\%, respectively. This performance aligns with previous observations that BP and EBP tend to classify fewer observations as outliers, reflecting their conservative approach. This cautious behavior is evident in the high precision values reported in Table~\ref{tab:random_sim}, with both BP and EBP achieving the highest precision. Many of the methods display an extremely high area under the curve (AUC), with kNN having the highest AUC for datasets with 1-10\%  and 10 to 20\% outliers. EBP has the second-highest AUC, while ECOD and DSVDD have the lowest AUCs.

\begin{table}[H]
\small
\caption{ Summary of measures for each method on data with a randomly generated number of modes from random distributions and random correlation structure.  Bold indicates the best performance. }
\label{tab:random_sim}
\begin{tabular}{@{}p{1.5cm}llllllllllll@{}}  \toprule
Measure & \% Out   & BP   & EBP  & ISO     & ECOD   & LOF     & kNN     & DSVDD  & LUN     & VAE     \\ \midrule
CC      & none     & \textbf{97.878} & 94.050 & 87.986  & 89.645 & 91.243  & 91.015  & 89.650 & 89.650  & 89.650  \\ 
      & 1 to 10  & \textbf{99.040} & 96.881 & 93.504  & 90.395 & 93.078  & 92.743  & 88.392 & 91.559  & 91.559  \\
        & 10 to 20 & 96.401 & 94.502 & \textbf{97.035 } & 91.060 & 96.646  & 96.505  & 85.982 & 95.499  & 95.482  \\ \midrule
DR      & 1 to 10  & 99.803 & 99.608 & \textbf{100.00} & 99.350 & \textbf{100.00} & \textbf{100.00} & 98.233 &\textbf{100.00} & \textbf{100.00} \\
        & 10 to 20 & 98.208 & 98.734 & 99.998  & 97.455 & 99.950  & \textbf{100.000} & 94.614 & 99.927  & 99.917  \\ \midrule
PREC    & 1 to 10  & \textbf{99.211 }& 97.233 & 93.456  & 90.822 & 92.967  & 92.622  & 89.789 & 91.422  & 91.422  \\
        & 10 to 20 & \textbf{98.076} & 95.505 & 96.879  & 92.941 & 96.524  & 96.326  & 90.214 & 95.334  & 95.325  \\ \midrule
AUC     & 1 to 10  & 99.680 & 99.966 & 99.934  & 83.660 & 99.850  & \textbf{99.978 } & 48.067 & 99.574  & 99.701  \\
        & 10 to 20 & 99.662 & 99.972 & 99.934  & 83.574 & 99.839  & \textbf{99.983} & 53.324 & 99.639  & 99.552 \\ \bottomrule
\end{tabular}

\end{table}

\section{Example Data Comparison}\label{sec:Example_data}

 \cite{campos2016evaluation} used semantically meaningful datasets to evaluate outlier detection methods. The original datasets include a labeled class that can be assumed to be rare, and thus treated as outliers—for instance, a class of sick patients within a population dominated by healthy individuals. The datasets prepared by \cite{campos2016evaluation} come from benchmark data commonly used in outlier detection research. To prepare these datasets, the authors sampled (where possible) the outlying class at different rates: 20\%, 10\%, 5\%, and 2\%. To minimize bias, 10 versions were created for each percentage of outliers for every original dataset. Note that not all datasets could be prepared with all four outlier percentages due to insufficient observations in the outlier class. The authors provided both normalized datasets (scaled to [0,1] with duplicates removed) and non-normalized versions (which include duplicate observations). To replicate a realistic scenario where data pre-processing has occurred, we used the normalized datasets without duplicates, resulting in a total of 1,700 datasets. The datasets are available at: https://www.dbs.ifi.lmu.de/research/outlier-evaluation/, and the attribute types, number of attributes, and sample sizes for each original dataset are summarized in Table~\ref{tab:semantic_data}. While it is difficult to determine whether the data are multi-modal, several datasets contain more than two classes, suggesting that some may indeed include more than a single mode. {Given the poor performance of OCP in Section~\ref{sec:syn_data}, we did not include it in the results below.}  

\begin{table}[H]
\centering
\caption{\label{tab:semantic_data}Summary of the original datasets used in . All attributes are numeric except for InternetAds which has 1555 binary attributes.}
\begin{tabular}{@{}llllllll@{}}
\toprule
Dataset          & $n$    & Outliers & Attributes & Dataset   & $n$    & Outliers & Attributes \\ \midrule
Annthyroid       & 7200 & 534  &     21  & InternetAds      & 3264 & 454     & 1555*       \\
Arrhythmia       & 450  & 206  &  259   &  PageBlocks       & 5473 & 560    & 10    \\
Cardiotocography & 2126 & 471  &     21 &  Parkinson        & 195  & 147    & 22       \\
HeartDisease     & 270  & 120  &     13 &   Pima             & 768  & 268     & 8        \\
Hepatitis        & 80   & 13      & 19  &  SpamBase         & 4601 & 1813   & 57      \\
Stamps           & 340  & 31     & 9 &   Wilt             & 4839 & 261     & 5     \\ \bottomrule
\end{tabular}

\end{table}

Table~\ref{tab:CC_semantic} shows the average CC across  all versions of the datasets. This includes datasets with 2-20\% outliers. EBP has the highest CC in 6 of the 11 datasets with BP having two of the highest and  ISO, kNN and VAE having a 1 of the highest. \text{red}{EBP has the highest average CC followed by ECOD and kNN.}

Table~\ref{tab:AUC_semantic} shows the average AUC across all versions of the data.  {ECOD has the highest average AUC followed by EBP. VAE has the third highest and ISO the fourth}. Since AUC is independent of the specific choice of cutoff indicating EBP is better at separating the outliers from the inliers for these datasets.

Although not shown in the main body of the paper, the DR is sporadic across the methods and the datasets with ISO and BP having 3 and 4 of the top DR, respectively (see the supplementary materials).  In general the DR for these datasets are low and variable. For example for the Hepatitis data, ISO had a detection rate of 85\% whereas BP had a detection rate of 0. Likewise ISO has a 0\% detection rate for the InternetAds data whereas LOF has 55\% DR. {It is not surprising that we observe poor performance on the InternetAds dataset, as it is composed of binary variables, and BP and EBP, in their current implementation, are not recommended for binary data. We suspect that the small size of the hepatitis dataset makes it more suitable for a method that is more sensitive, such as those using ensemble techniques. This inconsistency in performance aligns with findings in the literature on unsupervised outlier detection; for instance, see \cite{campos2016evaluation}.}

{In 4 out the 8 datasets EBP has the highest precision and the overall highest precision average.} In two cases BP has a precision of 0\% and in one case ISO has a precision of 0\%. Lastly Table~\ref{tab:time} shows the average processing time original version of each of the datasets. BP had three of the fastest processing times but kNN had the overall average fastest processing time. Processing times were measured in seconds using a workstation with an Intel Xeon Processor E5-2687W and dual SLI NVIDIA Quadro P5000 graphical processing units (GPU).

\begin{table}[H] 
\small
\caption{\label{tab:CC_semantic} Average correct classification rate overall of the semantically created normalized data sets without duplicates for each outlier detection method. }
\begin{tabular}{@{}l|lllllllll@{}} \toprule
Data            & BP   & EBP  & ISO    & ECOD   & LOF    & KNN    & DSVDD   & LUN    & VAE    \\ \midrule
Annthyroid       & 86.270 & \textbf{94.730} & 93.350 & 88.860 & 88.500 & 87.950 & 89.090 & 87.970 & 87.490 \\
Arrhythmia      & \textbf{91.566} & 90.668 & 91.273 & 89.924 & 89.216 & 89.755 & 87.021 & 89.533 & 86.970 \\
Cardiotocography & 84.090 & \textbf{89.770} & 87.070 & 87.330 & 84.920 & 85.890 & 83.630 & 85.180 & 86.670 \\
HeartDisease    & 83.930 & \textbf{91.210} & 70.300 & 86.740 & 83.140 & 85.330 & 84.070 & 84.050 & 86.700 \\
Hepatitis       &  \textbf{93.130} &91.380 & 60.820 & 85.890 & 86.580 & 84.930 & 84.500 & 88.230 & 88.360 \\
InternetAds     & 94.360 & \textbf{94.820} & 94.360 & 94.260 & 89.980 & 91.060 & 88.400 & 89.490 & 89.730 \\
PageBlock       & 84.310 & 87.040 & 89.770 & 90.700 & 90.290 & 90.790 & 88.840 & 90.620 & \textbf{90.880} \\
Parkinson       & 86.120 & 87.340 & 85.910 & 85.930 & 87.750 & \textbf{88.320} & 82.980 & 87.280 & 87.410 \\
Pima            & 80.900 & \textbf{85.790} & 82.980 & 85.050 & 83.610 & 84.500 & 82.050 & 84.730 & 84.470 \\
SpamBase         & 84.720 & 90.170 & \textbf{90.290} & 85.930 & 83.070 & 84.570 & 83.600 & 84.780 & 84.740 \\
Stamps          & 91.498 & \textbf{91.641} & 89.526 & 89.503 & 87.852 & 89.980& 87.003& 89.682 & 89.318 \\ \midrule
Average & 87.354 &	\textbf{90.414}	& 85.059 &	88.192 &	86.810 &	87.552 &	85.562 &	87.413 & 87.522
\\ \midrule
Rank & 6	&1	&9	&2	&7	&3	&8	&5	&4
\\ 
\bottomrule
\end{tabular}

\end{table}

\begin{table}[H]
\small
\caption{\label{tab:AUC_semantic} Average AUC rate overall of the semantically created normalized data sets without duplicates for each outlier detection methods.}
\begin{tabular}{@{}l|lllllllll@{}} \toprule
Data             & BP   & EBP  & ISO    & ECOD   & LOF    & KNN    & DSVDD   & LUN    & VAE    \\ \midrule
Annthyroid       & 50.363 & 64.337 & 67.155 & \textbf{76.828} & 68.060 & 64.714 & 72.420 & 64.886 & 60.396 \\
Arrhythmia       & 78.711 & 77.801 & 76.916 & 74.344 & \textbf{79.526} & \textbf{79.526}  & 60.853 & 74.172 & 77.055 \\
Cardiotocography & 78.540 & 79.396 & 75.904 & \textbf{82.877} & 68.570 & 66.534 & 50.653 & 64.641 & 79.333 \\
HeartDisease     & 77.012 & \textbf{84.692} & 80.960 & 79.623 & 69.895 & 76.111 & 64.573 & 70.068 & 79.628 \\
Hepatitis        & 74.534 & 78.077 & 76.269 & 78.689 & 78.124 & 73.838 & 58.600 & 80.860 & \textbf{81.073} \\
InternetAds      & 70.416 & 68.690 & 62.374 & 71.374 & 75.318 & \textbf{81.310} & 72.129 & 79.446 & 78.652 \\
PageBlock        & 91.307 & 86.824 & 91.220 & \textbf{92.873}  & 82.313 & 89.700 & 58.532 & 88.142 & 91.980  \\
Parkinson        & 74.233 & 80.435 & 81.938 & 75.069 & 76.631 & \textbf{84.010} & 59.369 & 85.894 & 72.850 \\
Pima             & 69.655 & 72.148 & 71.725 & 66.830 & 66.643 & \textbf{73.476} & 49.617 & 71.553 & 69.870 \\
SpamBase         & 67.949 & \textbf{77.375} & 77.054 & 75.787 & 68.342 & 72.187 & 57.597 & 72.120 & 71.742 \\
Stamps           & 90.839 & 89.166 & \textbf{91.632} & 90.827  & 90.673 & 91.606  & 57.704 & 85.398 & 91.515 \\ \midrule
Average      &   74.869 &	78.086 &	77.559 &	\textbf{78.647}	&74.918&	77.547	&60.186&	76.107	&77.645
\\ 
\midrule
Rank &8	&2	&4	&1	&7&	5&	9&	6&	3
\\ 
\bottomrule
\end{tabular}

\end{table}

\begin{table}[H]
\small
\caption{\label{tab:time} Average processing time in seconds for normalized data sets without duplicates. Processing times were measured in seconds using a workstation with an Intel Xeon Processor E5-2687W and dual SLI NVIDIA Quadro P5000 graphical processing units (GPU).}
\begin{tabular}{@{}l|lllllllll@{}} \toprule
Data            & BP   & EBP & ISO   & ECOD  & LOF    & KNN   & DSVDD   & LUN   & VAE     \\ \midrule
Annthyroid       & 11.475 & 1026.208 & \textbf{0.431} & 5.972 & 1.130  & 1.126 & 14.370 & 3.785 & 26.004  \\
Arrhythmia      & 0.119  & 3.346    & 0.277 & \textbf{0.043} & 0.159  & 0.160 & 2.404  & 2.543 & 4.946   \\
Cardiotocography & 0.821  & 99.658   & 0.316 & \textbf{0.262} & 0.382  & 0.382 & 5.627  & 3.660 & 150.287 \\
HeartDisease    & \textbf{0.017 } & 3.208    & 0.263 & 0.020 & 0.037  & 0.035 & 2.237  & 1.661 & 3.434   \\
Hepatitis       & \textbf{0.008}  & 0.973    & 0.258 & 0.021 & 0.220  & 0.220 & 1.973  & 3.054 & 2.835   \\
InternetAds      & 7.325  & 155.107  & 1.780 & 1.036 & \textbf{0.478}  & 0.688 & 10.339 & 4.824 & 925.483 \\
PageBlock       & 6.243  & 1379.576 & 0.383 & 3.662 & \textbf{0.116 } & 0.179 & 10.945 & 3.038 & 16.824  \\
Parkinson       & \textbf{0.007}  & 0.417    & 0.257 & 0.032 & 0.221  & 0.222 & 2.601  & 2.699 & 3.039   \\
Pima            & 0.094 & 15.080   & 0.281 & 0.143 & 27.987 & \textbf{0.037} & 3.042  & 1.677 & 4.810   \\
SpamBase         & 1.913  & 245.913  & \textbf{0.359 }& 1.169 & 0.598  & 0.621 & 7.002  & 4.084 & 14.090  \\
Stamps          & 0.066  & 7.804    & 0.268 & 0\textbf{.015} & 0.038  & 0.035 & 2.542  & 1.700 & 3.758   \\ \midrule
Average & 2.553&	267.026&	0.443&	1.125&	2.851&	\textbf{0.337}&	5.735&	2.975	&105.046
\\ 
\midrule
Rank & 4	& 9	& 2 &	3	&5&	1 &	7&	6&	8\\  
\bottomrule
\end{tabular} 

\end{table}

In the implementation of the methods in the PyOD package, the users must specify the percentage of outliers in the data. We set that to be 10\% for all of the methods.  In the supplementary materials, we summarize the DR, CC, AUC and precision for only the sampled datasets with 10\% outliers. Generally we see similar patterns as before. DR is fairly inconsistent with BP having three of the highest values. EBP dominates correct classification for the datasets with only 10\% outliers with the highest average. EBP had four of the highest AUC values followed by ISO with three of the highest values.  ISO shows a precision of 100\% for the Arrhythmia datasets with 10\% outliers, however, EBP has four of the highest precision values for the remainder of the datasets with 10\% outliers.


\section{Discussion}\label{sec:discussion}

We have introduced a novel method for outlier detection based on iteratively peeled observations using OCSVM and signed distances. Our method was compared with state-of-the-art approaches on both unimodal and multimodal synthetic data, as well as on semantically meaningful benchmark datasets. From the synthetic data studies, several conclusions can be drawn. When no outliers are present, BP achieves a higher correct classification rate than ISO in all but one case. All methods, including BP and EBP, demonstrate high CC and detection rates for both unimodal and bimodal data with 10\% outliers. LOF shows the highest average AUC for all synthetic data, with BP following closely. In the case of completely randomly generated synthetic data, BP has the highest CC rate for data with 0-10\% outliers and the highest precision. While kNN achieves the highest AUC, BP and EBP are comparable to the other methods in terms of AUC performance.
       
From the synthetic data comparison, we conclude that the performance of BP and EBP is on par with state-of-the-art methods when outliers are present, and superior when no outliers are present. For small sample sizes, preventing unnecessary data loss by retaining inliers is a significant advantage. {OCP demonstrates superior performance compared to other algorithms in scenarios without outliers, even in multimodal simulations, a situation for which it was not specifically designed. By estimating a single mean, OCP generates similar kernel distances for observations across different modes, as the mean generally falls between the modes. However, its performance deteriorates as expected in multimodal cases and when outliers are present. This often results in the misclassification of both inliers and outliers, making OCP the worst performer compared to other algorithms.}

The comparison of methods on the semantically created benchmark datasets leads to the following conclusions. EBP achieved the highest overall CC DR, indicating a conservative approach in identifying outliers. The method with the highest average DR was ISO, followed closely by VAE. Interestingly, VAE never achieved the highest DR for any single dataset but showed the most consistent performance across datasets. Although EBP did not have the highest DR, it did have the highest AUC and precision, suggesting that the default threshold for outlier identification may not be optimal. The most consistent and computationally efficient method was kNN. In contrast, EBP, in its current implementation, was the least efficient. However, BP demonstrated average computational efficiency compared to the other methods. Further work is needed to improve the computational efficiency of EBP, but if computational efficiency is critical, the BP method offers performance on par with other leading methods.

One weakness of BP and EBP lies in their handling of modes with significantly different sizes. In the synthetic data simulations with randomly generated modes, the modes varied in size but not drastically. If one mode contained the majority of the data while another was much smaller, BP and EBP would likely categorize the observations in the smaller mode as outliers. We tested the performance of BP and EBP against other algorithms in such scenarios. Although their CC rate was lower compared to cases where the modes were of similar sizes, BP and EBP still performed competitively and, in most cases, outperformed the competing methods.

BP and EBP were implemented with baseline parameters, and their performance could likely be enhanced through tuning, particularly by adjusting the outlier threshold, $h$. Further research is needed in this area. In the implementation of EBP, the parameter $c$ and the {randomly sampled} feature set, $\sqrt{p}$, were fixed at strict values. {Additionally, the weighting scheme applied to individual observations may not be optimal.} Adjusting these parameters could improve both the computational efficiency and sensitivity of the method. Lastly, EBP combines scores using a simple average, which may not be ideal. Future work could explore alternative methods for aggregating the results from each bagged set.

The time complexity of OCSVM using a standard quadratic programming solver is $\mathcal{O}(n^3)$ \citep{kang2019approximate}. The time complexity of BP depends on the number of OCSVM boundaries constructed, which in the worst case can be up to $n$ boundaries, making BP's worst-case complexity $\mathcal{O}(n^4)$. Although not implemented in this work, the number of boundaries can be reduced by adjusting the tuning parameters of OCSVM to construct smaller hyperspheres, allowing more observations to be peeled off in each iteration and thus requiring fewer boundaries. \cite{alam2020one} offer extensions to OCSVM that reduce time complexity, and these methods could be explored for improving BP's efficiency.

{Figure 1 in the supplementary materials shows the increase in processing time as the sample size increases for BP and this could be limiting for large data. Figure 2 shows that the processing time for BP only increases slightly with increasing dimension.  }

For both methods the boundaries created by OCSVM were set to be as large as possible and hence peel a small number of observations at a time.  For a dataset with a large number of observations, adjusting $q$ so that smaller boundaries are created each peel would improve the computational efficiency by requiring fewer peels.    


\bibliographystyle{apalike} 
\bibliography{ref.bib}

\end{document}